%% file: main.tex
\documentclass[10pt,twocolumn,letterpaper]{article}

\usepackage[pagenumbers]{main}

\usepackage[table,dvipsnames]{xcolor}
\usepackage{microtype}
\usepackage{graphicx}
\usepackage{booktabs}
\usepackage{amsmath}
\usepackage{amssymb}
\usepackage{mathtools}
\usepackage{amsthm}

\usepackage{color}
\usepackage{enumitem}
\usepackage{multirow}
\usepackage{makecell}

\usepackage{algorithmic}
\usepackage{algorithm}
\usepackage{etoolbox,siunitx}

\definecolor{lblue}{rgb}{0.21,0.49,0.74}

\usepackage[pagebackref=false,breaklinks,colorlinks,citecolor=Orchid]{hyperref}

\title{The 2nd Place Solution from the 3D Semantic Segmentation Track in the\\2024 Waymo Open Dataset Challenge}

\author{Qing Wu\\
Marvell Technology\\
{\tt\small qing20210419@gmail.com}
}

\begin{document}

\maketitle

\input{sections/0_abstract}

\input{sections/1_intro}
\input{sections/3_approach}
\input{sections/4_experiments}
\input{sections/5_conclusion}

{
\small
\bibliographystyle{ieeenat_fullname}
\bibliography{main}
}

\end{document}

%% file: sections/0_abstract.tex
\begin{abstract}
3D semantic segmentation is one of the most crucial tasks in driving perception. The ability of a learning-based model to accurately perceive dense 3D surroundings often ensures the safe operation of autonomous vehicles. However, existing LiDAR-based 3D semantic segmentation databases consist of sequentially acquired LiDAR scans that are long-tailed and lack training diversity. In this report, we introduce MixSeg3D, a sophisticated combination of the strong point cloud segmentation model with advanced 3D data mixing strategies. Specifically, our approach integrates the MinkUNet family with LaserMix and PolarMix, two scene-scale data augmentation methods that blend LiDAR point clouds along the ego-scene's inclination and azimuth directions. Through empirical experiments, we demonstrate the superiority of MixSeg3D over the baseline and prior arts. Our team achieved 2nd place in the 3D semantic segmentation track of the 2024 Waymo Open Dataset Challenge.
\end{abstract}

%% file: sections/1_intro.tex
\section{Introduction}
\label{sec:intro}

The rapid advancement of autonomous driving technology has underscored the critical importance of accurate and efficient perception systems \cite{geiger2012kitti}. Among the various tasks involved in driving perception, 3D semantic segmentation stands out as a pivotal component \cite{fong2022panoptic-nuScenes,sun2020waymoOpen,behley2019semanticKITTI,kong2024lasermix2,xu2024superflow}. 
It involves the classification of each point in a LiDAR point cloud into distinct semantic categories, such as vehicles, pedestrians, and road surfaces, thereby enabling the autonomous vehicle to understand and navigate its environment safely \cite{behley2021semanticKITTI,kong2024calib3d}.

LiDAR sensors have become a cornerstone in autonomous driving due to their ability to provide high-resolution 3D information about the surrounding environment \cite{caesar2020nuScenes}. Unlike cameras, which capture 2D images, LiDAR sensors emit laser pulses and measure the time it takes for the pulses to return after hitting an object \cite{milioto2019rangenet++,douillard2011lidarseg}. This process generates detailed 3D point clouds that represent the spatial structure of the environment \cite{wang2018pointseg,ando2023rangevit,2023CLIP2Scene,kong2023rethinking,liu2023seal}.

3D semantic segmentation of LiDAR point clouds is crucial for various autonomous driving tasks, including object detection, scene understanding, and navigation \cite{xu2021rpvnet,liu2023uniseg}. The challenge lies in accurately identifying and categorizing each point in the cloud, which requires sophisticated algorithms capable of handling large-scale data and diverse environmental conditions. Traditional methods often struggle with the complexity and variability of real-world driving scenarios, highlighting the need for advanced approaches \cite{hu2021sensatUrban}.

\begin{figure*}[t]
    \centering
    \begin{subfigure}[b]{0.49\textwidth}
        \centering
        \includegraphics[width=\textwidth]{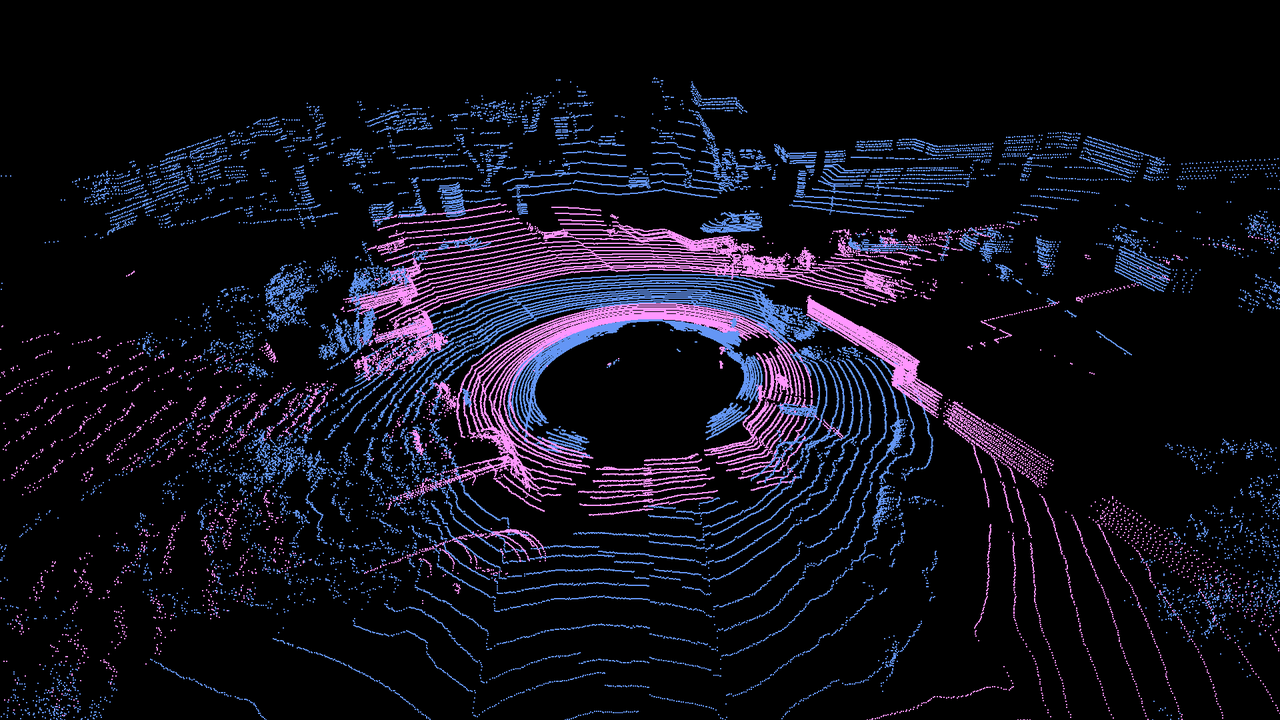}
        \caption{LaserMix}
        \label{fig:lasermix}
    \end{subfigure}
    \hfill
    \begin{subfigure}[b]{0.49\textwidth}
        \centering
        \includegraphics[width=\textwidth]{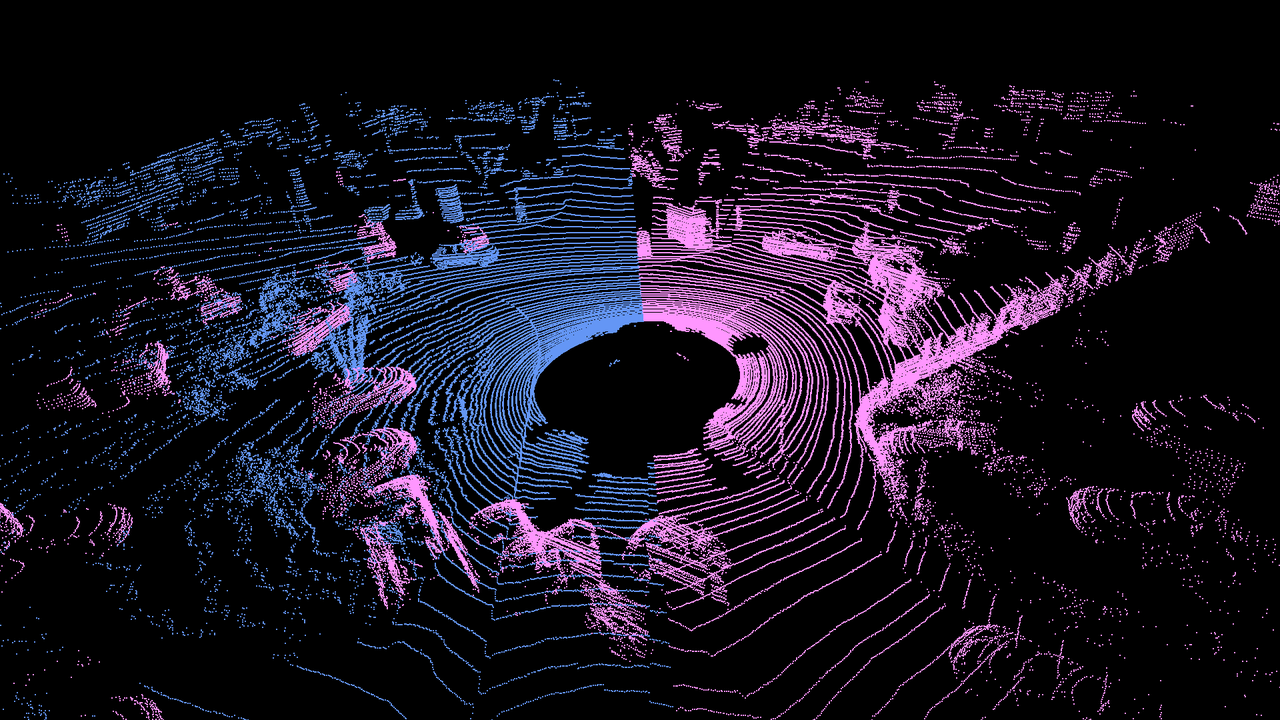}
        \caption{PolarMix}
        \label{fig:polarmix}
    \end{subfigure}
    \caption{Illustrative examples of LiDAR scene augmentations using the (a) LaserMix \cite{kong2023laserMix} and (b) PolarMix \cite{xiao2022polarmix} strategies.}
    \label{fig:teaser}
\end{figure*}

The 2024 Waymo Open Dataset Challenge \cite{sun2020waymoOpen} provides a competitive platform for exploring and advancing solutions in this domain. The challenge focuses on segmenting LiDAR point clouds into semantic classes, aiming to push the boundaries of current methodologies and foster innovation. This competition attracts top research teams worldwide, encouraging the development of novel approaches and the exchange of cutting-edge ideas.

Since the LiDAR scans are often sequentially acquired, the consecutive point clouds share overlapped class and spatial distributions, which might lead to sub-par performance. In this report, we present our approach, \textcolor{Orchid}{\textbf{MixSeg3D}}, which combines a strong point cloud segmentation model with advanced 3D data mixing strategies. Our method integrates the MinkUNet family \cite{choy2019minkowski} with LaserMix \cite{kong2023laserMix} and PolarMix \cite{xiao2022polarmix}, two novel scene-scale data augmentation techniques.
MinkUNet \cite{choy2019minkowski}, known for its efficient sparse convolutional operations, provides a robust backbone for our segmentation framework. LaserMix \cite{kong2023laserMix} and PolarMix \cite{xiao2022polarmix} blend LiDAR point clouds along the ego-scene's inclination and azimuth directions, respectively, enhancing the LiDAR segmentation model's ability to generalize across different environments \cite{kong2023robo3D,hahner2021fog}. 

By incorporating LaserMix \cite{kong2023laserMix} and PolarMix \cite{xiao2022polarmix}, we introduce additional variations in the training data, allowing the MinkUNet model to learn from a wider range of scenarios. This approach improves segmentation accuracy and enhances the model's robustness to domain shifts and out-of-distribution data.
Our empirical studies demonstrate that \textcolor{Orchid}{\textbf{MixSeg3D}} outperforms the baseline and prior methods, achieving superior segmentation accuracy. We achieved 69.83\% mIoU on the leaderboard, culminated in securing 2nd place in the 3D semantic segmentation track of the 2024 Waymo Open Dataset Challenge.

%% file: sections/3_approach.tex
\section{Approach}

Recent years have seen significant progress in the field of 3D semantic segmentation, driven by the development of powerful deep learning models and the availability of large annotated datasets. Neural networks, particularly convolutional neural networks and their 3D variants \cite{tang2020searching,yan2018second,tang2022torchsparse,tang2023torchsparse++}, have demonstrated remarkable capabilities in learning rich representations from point cloud data. Architectures such as MinkUNet \cite{choy2019minkowski} have set new benchmarks in 3D semantic segmentation by leveraging hierarchical feature extraction and efficient spatial representations.

\subsection{Revisiting MinkUNet}
MinkUNet \cite{choy2019minkowski} is a widely recognized architecture for 3D semantic segmentation due to its efficient sparse convolutional operations and robust hierarchical feature extraction. The architecture employs sparse tensors to handle the high dimensionality and sparsity of LiDAR point clouds, making it computationally efficient while maintaining high accuracy.

MinkUNet's core design revolves around the use of Minkowski Engine, which facilitates the computation of sparse convolutions and offers significant improvements in processing speed and memory usage. This allows MinkUNet to scale effectively with large datasets and dense point clouds. The network architecture consists of an encoder-decoder structure, where the encoder extracts multi-scale features through a series of sparse convolutional layers, and the decoder reconstructs the segmented output using transposed sparse convolutions.

In our approach, we adopt the MinkUNet as the backbone due to its proven effectiveness in handling large-scale 3D point clouds. Different from the original MinkUNet configuration, we used a larger backbone, dubbed \textbf{MinkUNet-101}, to learn more informative features from the complex point clouds in the Waymo Open dataset \cite{sun2020waymoOpen}. The hierarchical feature extraction capability of MinkUNet allows it to capture fine-grained details and global context, which are crucial for accurate semantic segmentation. For additional details, kindly refer to Choy~\etal \cite{choy2019minkowski}.

\subsection{MixSeg3D}
To further enhance the performance of MinkUNet, we introduce MixSeg3D, a sophisticated combination of the MinkUNet backbone with advanced 3D data mixing strategies: LaserMix \cite{kong2023laserMix} and PolarMix \cite{xiao2022polarmix}. These techniques are designed to augment the training data by blending LiDAR point clouds in innovative ways, thereby improving the model's generalization and robustness. \cref{fig:teaser} provides illustrative examples of conducting LaserMix and PolarMix on LiDAR point clouds, where the two colors represent points from two randomly sampled LiDAR scans.

\noindent\textbf{LaserMix} is a data augmentation technique that blends LiDAR point clouds along the ego-scene's inclination direction. By partitioning the point clouds and combining them at various angles, LaserMix generates diverse training samples that simulate different driving conditions and perspectives. This method helps the model learn to recognize objects and semantic classes from various inclinations, enhancing its ability to generalize to unseen data. We set the probability of conducting the LaserMix operation during training as $p_{1}$.

\noindent\textbf{PolarMix} complements LaserMix by blending LiDAR point clouds along the azimuth direction. This technique rotates the point clouds around the vertical axis, creating augmented samples that represent different orientations. PolarMix ensures that the model can handle variations in object orientation and spatial distribution, which are common in real-world driving scenarios. We set the probability of conducting the PolarMix operation during training as $p_{2}$.

\noindent\textbf{Test Time Augmentation (TTA)} is a technique used during model inference to enhance segmentation accuracy. This approach involves applying various transformations to the input data at test time and aggregating the predictions from multiple augmented versions of the data. In this competition, we adopted rotations, scaling, flipping, and shifting.

%% file: sections/4_experiments.tex
\section{Experiments}

\begin{table}[t]
\centering
\caption{The 3D semantic segmentation results from different participants in the 2024 Waymo Open Dataset Challenge.}
\resizebox{0.48\textwidth}{!}{
    \begin{tabular}{l|c|c}
    \toprule
    \textbf{Method} & \textbf{mIoU} & \textbf{Access Date}
    \\\midrule\midrule
    PTv3-EX & $72.76$ & $24/05/2024$, $12:08:05$
    \\
    \rowcolor{Orchid!10}\textcolor{Orchid}{\textbf{MixSeg3D}} & $69.83$ & $24/05/2024$, $10:48:19$
    \\
    RangeFormer++ & $69.06$ & $18/05/2024$, $22:53:51$
    \\
    PointTransformer-LR & $68.92$ & $24/05/2024$, $09:50:26$
    \\
    vFusedSeg3D & $68.79$ & $24/05/2024$, $03:07:11$
    \\
    vSeg3D & $68.06$ & $15/05/2024$, $22:19:17$
    \\
    SPVCNN++ & $68.05$ & $10/05/2024$, $03:40:55$
    \\
    \bottomrule
    \end{tabular}
}
\label{tab:comparative}
\end{table}

\begin{table*}[t]
\centering
\caption{The class-wise 3D semantic segmentation results of MixSeg3D from the 2024 Waymo Open Dataset Challenge leaderboard.}
\resizebox{\textwidth}{!}{
    \begin{tabular}{cccccccccccccccccccccc}
    \toprule
    \rotatebox{90}{Car} & \rotatebox{90}{Truck} & \rotatebox{90}{Bus} & \rotatebox{90}{Other Vehicle} & \rotatebox{90}{Motorcyclist} & \rotatebox{90}{Bicyclist} & \rotatebox{90}{Pedestrian} & \rotatebox{90}{Sign} & \rotatebox{90}{Traffic Light} & \rotatebox{90}{Pole} & \rotatebox{90}{Construction Cone~} & \rotatebox{90}{Bicycle} & \rotatebox{90}{Motorcycle} & \rotatebox{90}{Building} & \rotatebox{90}{Vegetation} & \rotatebox{90}{Tree Trunk} & \rotatebox{90}{Curb} & \rotatebox{90}{Road} & \rotatebox{90}{Lane Marker} & \rotatebox{90}{Other Ground} & \rotatebox{90}{Walkable} & \rotatebox{90}{Sidewalk}
    \\\midrule\midrule
    \rowcolor{Orchid!10}$95.6$ & $70.7$ & $73.5$ & $29.6$ & $8.4$ & $88.5$ & $91.5$ & $73.1$ & $32.2$ & $80.2$ & $60.0$ & $70.6$ & $80.3$ & $97.0$ & $86.8$ & $73.2$ & $75.1$ & $92.6$ & $48.5$ & $51.5$ & $70.6$ & $86.7$
    \\
    \bottomrule
    \end{tabular}
}
\label{tab:comparative_class}
\end{table*}

\begin{table}[t]
\centering
\caption{Ablation study on different training and evaluation configurations in MixSeg3D. Results reported are from the official validation set of the 2024 Waymo Open Dataset Challenge.}
\resizebox{0.48\textwidth}{!}{
    \begin{tabular}{l|c|c|c}
    \toprule
    \textbf{Backbone} & \textbf{Aug} & \textbf{TTA} & \textbf{mIoU}
    \\\midrule\midrule
    MinkUNet-18 & None & None & $68.02$
    \\
    MinkUNet-34 & None & None & $70.18$
    \\
    MinkUNet-50 & None & None & $70.34$
    \\
    \rowcolor{Orchid!10}MinkUNet-101 & None & None & $70.98$
    \\\midrule
    MinkUNet-101 & LaserMix & None & $71.37$
    \\
    MinkUNet-101 & PolarMix & None & $71.31$
    \\
    \rowcolor{Orchid!10}MinkUNet-101 & LaserMix + PolarMix & None & $72.06$
    \\\midrule
    MinkUNet-101 & LaserMix + PolarMix & $3$ times & $72.41$
    \\
    MinkUNet-101 & LaserMix + PolarMix & $6$ times & $72.67$
    \\
    \rowcolor{Orchid!10}MinkUNet-101 & LaserMix + PolarMix & $8$ times & $74.03$
    \\
    MinkUNet-101 & LaserMix + PolarMix & $10$ times & $73.67$
    \\
    \bottomrule
    \end{tabular}
}
\label{tab:ablation}
\end{table}

\subsection{Experimental Settings}
\noindent\textbf{Datasets.}
We follow the 2024 Waymo Open Dataset Challenge rules in preparing the training, validation, and test data. Specifically, the model was trained on the official \textit{train} split of Waymo Open \cite{sun2020waymoOpen} and tested on the associated \textit{val} and \textit{test} splits. The number of samples from each of the three splits is $\sim$24,000, $\sim$7,000, and $\sim$3,000, respectively.

\noindent\textbf{Implementation Details.}
Our model is implemented based on MMDetection3D \cite{mmdet3d2020}. Our model was trained with an effective batch size of 32 and a learning rate of 0.008 across 4 NVIDIA A100 GPUs. We adopted the AdamW optimizer and OneCycle learning rate scheduler for model optimization \cite{loshchilov2018adamw}. To enhance training data diversity, we used random rotations, scaling, flipping, and shifting, before the LaserMix and PolarMix operations. After that, we set the probabilities of conducting LaserMix and PolarMix, \ie, $p_1$ and $p_2$, as $0.8$ and $1.0$, respectively. After training, we set a combo of $8$ times of TTA during the evaluation. The model was trained for 80 epochs on the official training set and then tested on the corresponding validation and test sets.

\noindent\textbf{Evaluation Protocol.}
Following the official evaluation protocol of the 2024 Waymo Open Dataset Challenge, we report the Intersection-over-Union (IoU) scores for each class and the mean IoU (mIoU) scores across all semantic classes.

\subsection{Experimental Results}
\cref{tab:comparative} presents the mIoU scores of different teams in this year's challenge. Our proposed MixSeg3D achieved a competitive mIoU score of 69.83\%, securing second place among the participants. This performance is notable considering the diversity and complexity of the Waymo Open dataset, indicating the robustness and effectiveness of our approach.

Furthermore, \cref{tab:comparative_class} details the class-wise IoU scores for MixSeg3D. The results demonstrate that MixSeg3D performs particularly well in several key classes such as Car (95.6\%), Pedestrian (91.5\%), and Building (97.0\%). These high IoU scores in critical classes underscore the model's ability to accurately segment important objects in the driving environment, which is essential for driving perception.

Despite the strong overall performance, there are areas where MixSeg3D shows room for improvement. For instance, the IoU scores for classes such as Motorcyclist (8.4\%) and Traffic Light (32.2\%) indicate challenges in segmenting smaller or less frequent objects. These observations highlight potential directions for future enhancements, such as incorporating more targeted data augmentation techniques or refining the model architecture to better capture these classes. 

To understand the efficacy of each of the backbones and data augmentation techniques applied, we conduct an ablation study and show the results in \cref{tab:ablation}. We first study the model capacity. We found that larger backbone networks (\eg, MinkUNet-50 and MinkUNet-101, which contain 31.9M and 70.8M parameters, respectively) tend to yield higher segmentation accuracy, as the larger sets of parameters often learn more meaningful representations. We then study the efficacy of LaserMix, PolarMix, and TTA. We found the best possible configuration by combining all of them together and setting proper hyperparameters.

It is worth mentioning that the LaserMix and PolarMix operations are very efficient in practice. During the model training, these two techniques manipulate the LiDAR point clouds ``on-the-fly'' at the point level, which can be regarded as basic tensor operations, such as adding, removing, replacing, \etc. However, the time consumption of applying TTA is notably longer than plain evaluation. This is because the TTA operation requires multiple times of model inference, and, to a certain extent, is not very practical in terms of actual deployment. 

Overall, the experimental results validate the effectiveness of MixSeg3D. The integration of MinkUNet with advanced data augmentation techniques, LaserMix and PolarMix, has proven to be a successful strategy for enhancing 3D semantic segmentation performance.

%% file: sections/5_conclusion.tex
\section{Conclusion}

We presented MixSeg3D, a novel approach for 3D semantic segmentation that integrates the MinkUNet backbone with advanced data augmentation techniques, LaserMix and PolarMix. Our method significantly improves 3D segmentation accuracy and robustness in diverse driving environments, achieving a competitive mIoU score of 69.83\%, securing second place. These results validate our approach and demonstrate its potential for advancing 3D semantic segmentation in autonomous driving.

%% file: main.bbl
\begin{thebibliography}{30}
\providecommand{\natexlab}[1]{#1}
\providecommand{\url}[1]{\texttt{#1}}
\expandafter\ifx\csname urlstyle\endcsname\relax
  \providecommand{\doi}[1]{doi: #1}\else
  \providecommand{\doi}{doi: \begingroup \urlstyle{rm}\Url}\fi

\bibitem[Ando et~al.(2023)Ando, Gidaris, Bursuc, Puy, Boulch, and Marlet]{ando2023rangevit}
Angelika Ando, Spyros Gidaris, Andrei Bursuc, Gilles Puy, Alexandre Boulch, and Renaud Marlet.
\newblock Rangevit: Towards vision transformers for 3d semantic segmentation in autonomous driving.
\newblock In \emph{IEEE/CVF Conference on Computer Vision and Pattern Recognition}, pages 5240--5250, 2023.

\bibitem[Behley et~al.(2019)Behley, Garbade, Milioto, Quenzel, Behnke, Stachniss, and Gall]{behley2019semanticKITTI}
Jens Behley, Martin Garbade, Andres Milioto, Jan Quenzel, Sven Behnke, Cyrill Stachniss, and Juergen Gall.
\newblock Semantickitti: A dataset for semantic scene understanding of lidar sequences.
\newblock In \emph{IEEE/CVF International Conference on Computer Vision}, pages 9297--9307, 2019.

\bibitem[Behley et~al.(2021)Behley, Garbade, Milioto, Quenzel, Behnke, Gall, and Stachniss]{behley2021semanticKITTI}
Jens Behley, Martin Garbade, Andres Milioto, Jan Quenzel, Sven Behnke, Jürgen Gall, and Cyrill Stachniss.
\newblock Towards 3d lidar-based semantic scene understanding of 3d point cloud sequences: The semantickitti dataset.
\newblock \emph{International Journal of Robotics Research}, 40:\penalty0 959--96, 2021.

\bibitem[Caesar et~al.(2020)Caesar, Bankiti, Lang, Vora, Liong, Xu, Krishnan, Pan, Baldan, and Beijbom]{caesar2020nuScenes}
Holger Caesar, Varun Bankiti, Alex~H Lang, Sourabh Vora, Venice~Erin Liong, Qiang Xu, Anush Krishnan, Yu Pan, Giancarlo Baldan, and Oscar Beijbom.
\newblock nuscenes: A multimodal dataset for autonomous driving.
\newblock In \emph{IEEE/CVF Conference on Computer Vision and Pattern Recognition}, pages 11621--11631, 2020.

\bibitem[Chen et~al.(2023)Chen, Liu, Kong, Zhu, Ma, Li, Hou, Qiao, and Wang]{2023CLIP2Scene}
Runnan Chen, Youquan Liu, Lingdong Kong, Xinge Zhu, Yuexin Ma, Yikang Li, Yuenan Hou, Yu Qiao, and Wenping Wang.
\newblock Clip2scene: Towards label-efficient 3d scene understanding by clip.
\newblock In \emph{IEEE/CVF Conference on Computer Vision and Pattern Recognition}, pages 7020--7030, 2023.

\bibitem[Choy et~al.(2019)Choy, Gwak, and Savarese]{choy2019minkowski}
Christopher Choy, JunYoung Gwak, and Silvio Savarese.
\newblock 4d spatio-temporal convnets: Minkowski convolutional neural networks.
\newblock In \emph{IEEE/CVF Conference on Computer Vision and Pattern Recognition}, pages 3075--3084, 2019.

\bibitem[Contributors(2020)]{mmdet3d2020}
MMDetection3D Contributors.
\newblock {MMDetection3D: OpenMMLab} next-generation platform for general {3D} object detection.
\newblock \url{https://github.com/open-mmlab/mmdetection3d}, 2020.

\bibitem[Douillard et~al.(2011)Douillard, Underwood, Kuntz, Vlaskine, Quadros, Morton, and Frenkel]{douillard2011lidarseg}
Bertrand Douillard, James Underwood, Noah Kuntz, Vsevolod Vlaskine, Alastair Quadros, Peter Morton, and Alon Frenkel.
\newblock On the segmentation of 3d lidar point clouds.
\newblock In \emph{IEEE International Conference on Robotics and Automation}, pages 2798--2805, 2011.

\bibitem[Fong et~al.(2022)Fong, Mohan, Hurtado, Zhou, Caesar, Beijbom, and Valada]{fong2022panoptic-nuScenes}
Whye~Kit Fong, Rohit Mohan, Juana~Valeria Hurtado, Lubing Zhou, Holger Caesar, Oscar Beijbom, and Abhinav Valada.
\newblock Panoptic nuscenes: A large-scale benchmark for lidar panoptic segmentation and tracking.
\newblock \emph{IEEE Robotics and Automation Letters}, 7\penalty0 (2):\penalty0 3795--3802, 2022.

\bibitem[Geiger et~al.(2012)Geiger, Lenz, and Urtasun]{geiger2012kitti}
Andreas Geiger, Philip Lenz, and Raquel Urtasun.
\newblock Are we ready for autonomous driving? the kitti vision benchmark suite.
\newblock In \emph{IEEE/CVF Conference on Computer Vision and Pattern Recognition}, pages 3354--3361, 2012.

\bibitem[Hahner et~al.(2021)Hahner, Sakaridis, Dai, and Gool]{hahner2021fog}
Martin Hahner, Christos Sakaridis, Dengxin Dai, and Luc~Van Gool.
\newblock Fog simulation on real lidar point clouds for 3d object detection in adverse weather.
\newblock In \emph{IEEE/CVF International Conference on Computer Vision}, pages 15283--15292, 2021.

\bibitem[Hu et~al.(2021)Hu, Yang, Khalid, Xiao, Trigoni, and Markham]{hu2021sensatUrban}
Qingyong Hu, Bo Yang, Sheikh Khalid, Wen Xiao, Niki Trigoni, and Andrew Markham.
\newblock Towards semantic segmentation of urban-scale 3d point clouds: A dataset, benchmarks and challenges.
\newblock In \emph{IEEE/CVF Conference on Computer Vision and Pattern Recognition}, pages 4977--4987, 2021.

\bibitem[Kong et~al.(2023{\natexlab{a}})Kong, Liu, Chen, Ma, Zhu, Li, Hou, Qiao, and Liu]{kong2023rethinking}
Lingdong Kong, Youquan Liu, Runnan Chen, Yuexin Ma, Xinge Zhu, Yikang Li, Yuenan Hou, Yu Qiao, and Ziwei Liu.
\newblock Rethinking range view representation for lidar segmentation.
\newblock In \emph{IEEE/CVF International Conference on Computer Vision}, pages 228--240, 2023{\natexlab{a}}.

\bibitem[Kong et~al.(2023{\natexlab{b}})Kong, Liu, Li, Chen, Zhang, Ren, Pan, Chen, and Liu]{kong2023robo3D}
Lingdong Kong, Youquan Liu, Xin Li, Runnan Chen, Wenwei Zhang, Jiawei Ren, Liang Pan, Kai Chen, and Ziwei Liu.
\newblock Robo3d: Towards robust and reliable 3d perception against corruptions.
\newblock In \emph{IEEE/CVF International Conference on Computer Vision}, pages 19994--20006, 2023{\natexlab{b}}.

\bibitem[Kong et~al.(2023{\natexlab{c}})Kong, Ren, Pan, and Liu]{kong2023laserMix}
Lingdong Kong, Jiawei Ren, Liang Pan, and Ziwei Liu.
\newblock Lasermix for semi-supervised lidar semantic segmentation.
\newblock In \emph{IEEE/CVF Conference on Computer Vision and Pattern Recognition}, pages 21705--21715, 2023{\natexlab{c}}.

\bibitem[Kong et~al.(2024{\natexlab{a}})Kong, Xu, Cen, Zhang, Pan, Chen, and Liu]{kong2024calib3d}
Lingdong Kong, Xiang Xu, Jun Cen, Wenwei Zhang, Liang Pan, Kai Chen, and Ziwei Liu.
\newblock Calib3d: Calibrating model preferences for reliable 3d scene understanding.
\newblock \emph{arXiv preprint arXiv:2403.17010}, 2024{\natexlab{a}}.

\bibitem[Kong et~al.(2024{\natexlab{b}})Kong, Xu, Ren, Zhang, Pan, Chen, Ooi, and Liu]{kong2024lasermix2}
Lingdong Kong, Xiang Xu, Jiawei Ren, Wenwei Zhang, Liang Pan, Kai Chen, Wei~Tsang Ooi, and Ziwei Liu.
\newblock Multi-modal data-efficient 3d scene understanding for autonomous driving.
\newblock \emph{arXiv preprint arXiv:2405.05258}, 2024{\natexlab{b}}.

\bibitem[Liu et~al.(2023{\natexlab{a}})Liu, Chen, Li, Kong, Yang, Xia, Bai, Zhu, Ma, Li, Qiao, and Hou]{liu2023uniseg}
Youquan Liu, Runnan Chen, Xin Li, Lingdong Kong, Yuchen Yang, Zhaoyang Xia, Yeqi Bai, Xinge Zhu, Yuexin Ma, Yikang Li, Yu Qiao, and Yuenan Hou.
\newblock Uniseg: A unified multi-modal lidar segmentation network and the openpcseg codebase.
\newblock In \emph{IEEE/CVF International Conference on Computer Vision}, pages 21662--21673, 2023{\natexlab{a}}.

\bibitem[Liu et~al.(2023{\natexlab{b}})Liu, Kong, Cen, Chen, Zhang, Pan, Chen, and Liu]{liu2023seal}
Youquan Liu, Lingdong Kong, Jun Cen, Runnan Chen, Wenwei Zhang, Liang Pan, Kai Chen, and Ziwei Liu.
\newblock Segment any point cloud sequences by distilling vision foundation models.
\newblock In \emph{Advances in Neural Information Processing Systems}, 2023{\natexlab{b}}.

\bibitem[Loshchilov and Hutter(2018)]{loshchilov2018adamw}
Ilya Loshchilov and Frank Hutter.
\newblock Decoupled weight decay regularization.
\newblock In \emph{International Conference on Learning Representations}, 2018.

\bibitem[Milioto et~al.(2019)Milioto, Vizzo, Behley, and Stachniss]{milioto2019rangenet++}
Andres Milioto, Ignacio Vizzo, Jens Behley, and Cyrill Stachniss.
\newblock Rangenet++: Fast and accurate lidar semantic segmentation.
\newblock In \emph{IEEE/RSJ International Conference on Intelligent Robots and Systems}, pages 4213--4220, 2019.

\bibitem[Sun et~al.(2020)Sun, Kretzschmar, Dotiwalla, Chouard, Patnaik, Tsui, Guo, Zhou, Chai, Caine, Vasudevan, Han, Ngiam, Zhao, Timofeev, Ettinger, Krivokon, Gao, Joshi, Zhang, Shlens, Chen, and Anguelov]{sun2020waymoOpen}
Pei Sun, Henrik Kretzschmar, Xerxes Dotiwalla, Aurelien Chouard, Vijaysai Patnaik, Paul Tsui, James Guo, Yin Zhou, Yuning Chai, Benjamin Caine, Vijay Vasudevan, Wei Han, Jiquan Ngiam, Hang Zhao, Aleksei Timofeev, Scott Ettinger, Maxim Krivokon, Amy Gao, Aditya Joshi, Yu Zhang, Jonathon Shlens, Zhifeng Chen, and Dragomir Anguelov.
\newblock Scalability in perception for autonomous driving: Waymo open dataset.
\newblock In \emph{IEEE/CVF Conference on Computer Vision and Pattern Recognition}, pages 2446--2454, 2020.

\bibitem[Tang et~al.(2020)Tang, Liu, Zhao, Lin, Lin, Wang, and Han]{tang2020searching}
Haotian Tang, Zhijian Liu, Shengyu Zhao, Yujun Lin, Ji Lin, Hanrui Wang, and Song Han.
\newblock Searching efficient 3d architectures with sparse point-voxel convolution.
\newblock In \emph{European Conference on Computer Vision}, pages 685--702, 2020.

\bibitem[Tang et~al.(2022)Tang, Liu, Li, Lin, and Han]{tang2022torchsparse}
Haotian Tang, Zhijian Liu, Xiuyu Li, Yujun Lin, and Song Han.
\newblock Torchsparse: Efficient point cloud inference engine.
\newblock \emph{Proceedings of Machine Learning and Systems}, 4:\penalty0 302--315, 2022.

\bibitem[Tang et~al.(2023)Tang, Yang, Liu, Hong, Yu, Li, Dai, Wang, and Han]{tang2023torchsparse++}
Haotian Tang, Shang Yang, Zhijian Liu, Ke Hong, Zhongming Yu, Xiuyu Li, Guohao Dai, Yu Wang, and Song Han.
\newblock Torchsparse++: Efficient point cloud engine.
\newblock In \emph{IEEE/CVF Conference on Computer Vision and Pattern Recognition Workshops}, pages 202--209, 2023.

\bibitem[Wang et~al.(2018)Wang, Shi, Yun, Tai, and Liu]{wang2018pointseg}
Yuan Wang, Tianyue Shi, Peng Yun, Lei Tai, and Ming Liu.
\newblock Pointseg: Real-time semantic segmentation based on 3d lidar point cloud.
\newblock \emph{arXiv preprint arXiv:1807.06288}, 2018.

\bibitem[Xiao et~al.(2022)Xiao, Huang, Guan, Cui, Lu, and Shao]{xiao2022polarmix}
Aoran Xiao, Jiaxing Huang, Dayan Guan, Kaiwen Cui, Shijian Lu, and Ling Shao.
\newblock Polarmix: A general data augmentation technique for lidar point clouds.
\newblock In \emph{Advances in Neural Information Processing Systems}, pages 11035--11048, 2022.

\bibitem[Xu et~al.(2021)Xu, Zhang, Dou, Zhu, Sun, and Pu]{xu2021rpvnet}
Jianyun Xu, Ruixiang Zhang, Jian Dou, Yushi Zhu, Jie Sun, and Shiliang Pu.
\newblock Rpvnet: A deep and efficient range-point-voxel fusion network for lidar point cloud segmentation.
\newblock In \emph{IEEE/CVF International Conference on Computer Vision}, pages 16024--16033, 2021.

\bibitem[Xu et~al.(2024)Xu, Kong, Shuai, Zhang, Pan, Chen, Liu, and Liu]{xu2024superflow}
Xiang Xu, Lingdong Kong, Hui Shuai, Wenwei Zhang, Liang Pan, Kai Chen, Ziwei Liu, and Qingshan Liu.
\newblock 4d contrastive superflows are dense 3d representation learners.
\newblock In \emph{European Conference on Computer Vision}, pages 58--80, 2024.

\bibitem[Yan et~al.(2018)Yan, Mao, and Li]{yan2018second}
Yan Yan, Yuxing Mao, and Bo Li.
\newblock Second: Sparsely embedded convolutional detection.
\newblock \emph{Sensors}, 18\penalty0 (10):\penalty0 3337, 2018.

\end{thebibliography}
